\documentclass{article}

\usepackage{PRIMEarxiv}

\usepackage{graphicx}
\usepackage{amsmath}
\usepackage{amssymb}
\usepackage{rotating}

\usepackage[utf8]{inputenc} 
\usepackage[T1]{fontenc}    
\usepackage{hyperref}       
\usepackage{url}            
\usepackage{booktabs}       
\usepackage{amsfonts}       
\usepackage{nicefrac}       
\usepackage{microtype}      
\usepackage{lipsum}
\usepackage{fancyhdr}       
\usepackage{graphicx}       
\graphicspath{{media/}}     

\usepackage[symbol]{footmisc}

\title{Design-time Fashion Popularity Forecasting in VR Environments}

\author{
Stefanos-Iordanis Papadopoulos\\
{\tt\small stefpapad@iti.gr}
\and
Christos Koutlis\\
{\tt\small ckoutlis@iti.gr}
\and
Anastasios Papazoglou-Chalikias\\
{\tt\small tpapazoglou@iti.gr}
\and
Symeon Papadopoulos\\
{\tt\small papadop@iti.gr}
\and
Spiros Nikolopoulos\\
{\tt\small nikolopo@iti.gr}\\
\\\hspace{-5cm}CERTH-ITI, Thessaloniki Greece\\
}

\begin{document}

\maketitle

\begin{center}
    \centering
    \includegraphics[width=\textwidth]{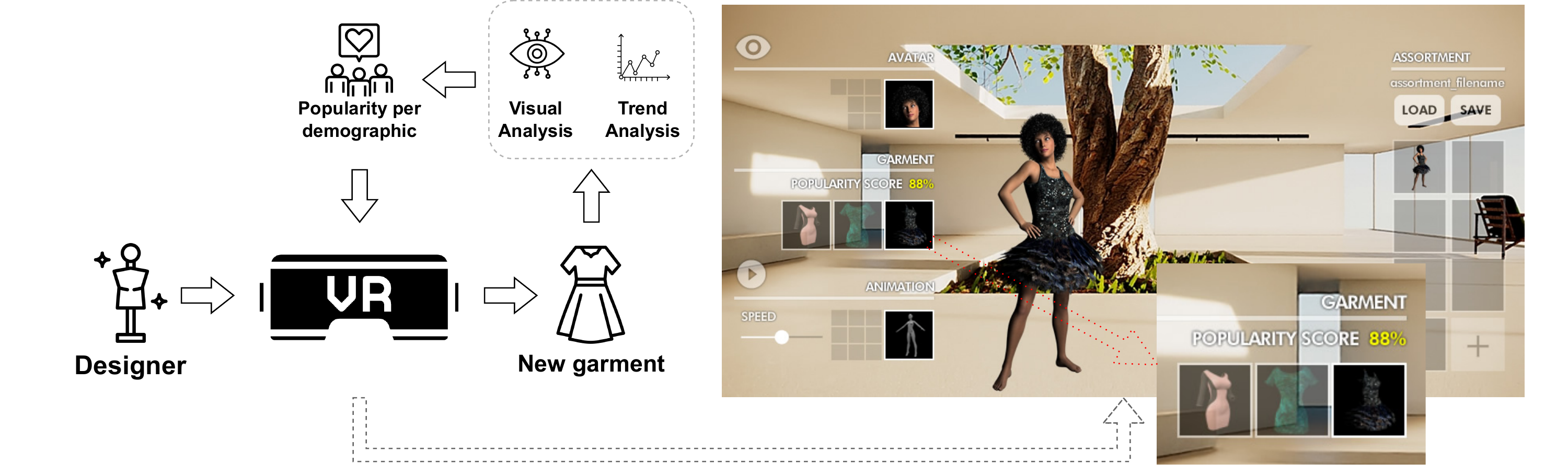}
    (Left) High level workflow of the VR designer application. The designer creates a new garment within the VR application.
    The proposed popularity forecasting model takes into account the visual features of the garment and the trends around its category and attributes and predicts the popularity for a given demographic group.
    (Right) The popularity score is presented within the user interface of the VR application and the designer may apply changes based on the feedback.
    \label{fig:teaser}
\end{center}%

\hspace{1cm}

\begin{abstract}
\footnotetext{This work has been presented at CVPR's Second Workshop on Ethical Considerations in Creative applications of Computer Vision, 2022.}
Being able to forecast the popularity of new garment designs is very important in an industry as fast paced as fashion, both in terms of profitability and reducing the problem of unsold inventory.
Here, we attempt to address this task in order to provide informative forecasts to fashion designers within a virtual reality designer application that will allow them to fine tune their creations based on current consumer preferences within an interactive and immersive environment.
To achieve this we have to deal with the following central challenges: (1) the proposed method should not hinder the creative process and thus it has to rely only on the garment's visual characteristics, 
(2) the new garment lacks historical data from which to extrapolate their future popularity and (3) fashion trends in general are highly dynamical.
To this end, we develop a computer vision pipeline fine tuned on fashion imagery in order to extract relevant visual features along with the category and attributes of the garment. We propose a hierarchical label sharing (HLS) pipeline for automatically capturing hierarchical relations among fashion categories and attributes.
Moreover, we propose MuQAR, a Multimodal Quasi-AutoRegressive neural network that forecasts the popularity of new garments by combining their visual features and categorical features while an autoregressive neural network is modelling the popularity time series of the garment's category and attributes. 
Both the proposed HLS and MuQAR prove capable of surpassing the current state-of-the-art in key benchmark datasets, DeepFashion for image classification and VISUELLE for new garment sales forecasting. 
\end{abstract}


\section{Introduction}
\label{sec:intro}

The ability to foresee the emergence and forecast the duration of trends in fashion is not only useful for individual consumers who want to be up-to-date with current trends but also for fashion brands and designers.
Accurate forecasting can help optimize production cycles and fine-tune the design of garments so that customers will more likely find appealing when they hit the shelves.
Furthermore, it could mitigate the problem of unsold inventory which is caused by a mismatch between supply and demand \cite{ekambaram2020attention} and has a significant environmental impact, with million tonnes of garments ending up in landfills or being burned \cite{niinimaki2020environmental}. 

Given this context, we propose a novel neural network architecture that forecasts the popularity of new garment designs based on their visual features. 
The popularity predictor is integrated within an interactive VR designer application, where designers receive an estimate of how popular their newly designed garments will be for a predefined market segment and date. 
Our goal is to aid the creative process and offer designers the option to fine tune their creations based on current consumer preferences. 
Modern fashion designers are already using 3D programs but VR is gradually gaining momentum to become an integrated part of the design process \cite{yang2021classifying}. 
The reason is that VR provides an immersive experience and allows the designer to simulate, examine and interact with their garments on 3D avatars before creating physical prototypes. 

For our approach to be interactive, supplementary and to not impede the creative process by requiring additional inputs from the designer (e.g written descriptions) we rely on computer vision; since fashion is a primarily visual-driven domain \cite{cheng2021fashion}. 
Accordingly, the designing process is expressed visually since it begins with sketching which communicates the shape, proportions, silhouette, fitment and builds up to being the blueprint of the garment by selecting fabrics and colors. 
Thus, we first develop a two-stage deep learning pipeline for detecting garments in an image and then classifying their fashion category (e.g shirt) and fine-grained attributes (e.g striped and collarless). We also propose ``hierarchical label sharing'' for learning hierarchical relations between fashion categories and attributes. 
Secondly, we propose MuQAR, a multimodal quasi-autoregressive method, for new garment popularity forecasting (NGPF); new garments that do not have past data. Both methods prove capable of surpassing the current state-of-the-art (SotA) in key benchmark datasets, DeepFashion \cite{liu2016deepfashion} and VISUELLE \cite{skenderi2021well} for image classification and forecasting respectively. 

\section{Related Work}
\label{sec:rw}
Recently, an increased research interest within the domain of computer vision towards fashion has been observed, 
addressing attribute classification \cite{papadopoulos2022attentive}, landmark detection \cite{liu2016deepfashion}, outfit compatibility \cite{papadopoulos2022victor}, fashion trends detection and garments or complete outfits popularity forecasting \cite{cheng2021fashion}. 
Computer vision models are used to identify fashion styles and attributes in order to discover trends in fashion \cite{al2017fashion, mall2019geostyle} or examine how trends spread among cities \cite{al2020paris}. 
Such studies result in interesting coarse level insights but are not particularly useful when applied to individual garments. For example, ``varsity jackets'' may be found to be trending but all individual ``varsity jackets'' would receive the same popularity score regardless of their unique aspects.
Autoregressive (AR) networks have been used to forecast the  popularity of individual garments based on their visual appearance \cite{lo2019dressing}. 
However, a newly designed garment, by definition, does not have past data, thus AR models can not be utilized.

Few works have focused on new garment popularity forecasting (NGPF).
In \cite{craparotta2019siamese} 
an image-based siamese neural network is proposed to identify similar items and infer their popularity to the new garment.
An AR recurrent neural network was proposed by \cite{ekambaram2020attention} that combines multimodality (images and text) with exogenous time series (holidays, discount season and events). 
The authors prepend two starting delimiters - since new products do not have past data - and then employ teacher enforcing for the proceeding steps during the training process. \cite{skenderi2021well} was critical of this approach because AR would suffer from first-step errors. Instead, they propose a non-AR transformer architecture that utilizes multimodality among images, categorical labels and time series of categorical labels collected from Google trends\footnote{\url{https://trends.google.com}}. 

Image-based NGPF is a new task and remains an open research challenge. One limitation of the aforementioned works is that they use computer vision networks pre-trained on ImageNet, a general purpose dataset, and therefore may not be able to recognize the intricacies and fine-grained aspects found in fashion imagery. For this reason we develop a computer vision pipeline, fine tune it on fashion imagery and then use it for forecasting. 
Additionally, there has been an increased interest for interactive VR-based collaborative environment for fashion design \cite{yang2021classifying}. However, to the best of our knowledge, our work is the first to integrate automatic NGPF in a VR designer application.

\label{sec:methodology}

\begin{figure*}[!ht]
    \centering
    \includegraphics[width=0.99\textwidth]{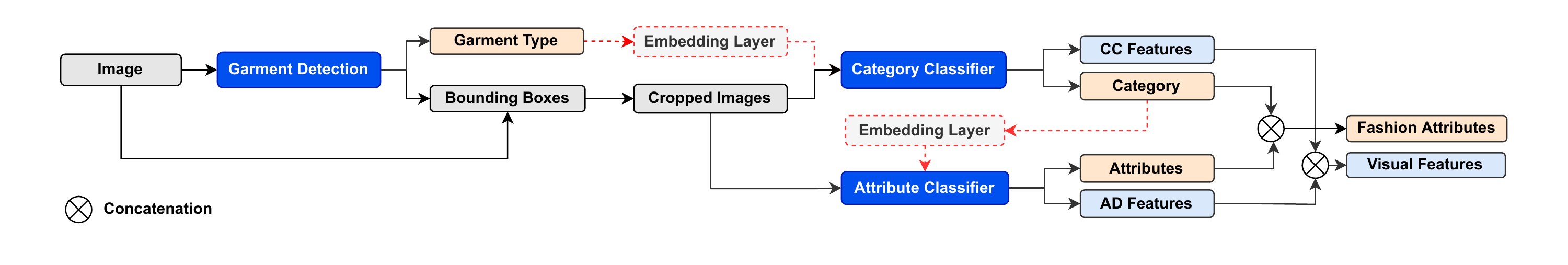}
    \caption{Workflow of the computer vision network. The red intermittent lines is optional and only used for HLS.}
    \label{fig:hls}
\end{figure*}

\section{Methodology}

\subsection{Pattern recognition on fashion imagery}
\label{subsec:cv}

To forecast the popularity of new garment designs based on their visual features, we must first train a computer vision network to extract those features and recognise relevant patterns in fashion imagery.  
To this end, we utilize transfer learning for fine-tuning pre-trained computer vision models on three tasks: 1) garment detection (GD), 2) category classification (CC) and 3) attributes classification (AC). 

For GD we fine-tune object detection models to identify the location of garments and their high-level type (upper-body, lower-body, full-body, footwear). GD allows inference on fashion imagery that depict complete outfits and enable CC and AC models to focus on individual garments.
Additionally, defining only four distinct classes for GD renders the task easier, reducing misclassification errors while focusing more on correct localisation. 
We experiment with four object detection architectures, namely Faster-RCNN \cite{ren2015faster}, CenterNet \cite{duan2019centernet} and EfficientDet-D1 and D2 \cite{tan2020efficientdet}. 
\\ \indent
For CC and AC we follow a conventional image classification transfer learning workflow where convolutional-based neural networks pre-trained on ImageNet are fine-tuned on a new dataset. We use the EfficientNet-B4 architecture pre-trained with the noisy students method on ImageNet \cite{xie2020self}. We unfreeze a portion of the EfficientNet-B4's layers and add a fully connected classification layer on top. 
We use image augmentation (horizontally flips, random rotation, random zooms by \textpm 10\%) for regularization.
Since CC is a multi-class problem we use softmax for the classification layer and the categorical cross entropy loss. AC is a multi-label problem so we use the sigmoid activation function and the binary cross entropy respectively. 
\\ \indent
Furthermore, we make use of hierarchical label sharing (HLS), a simple and lightweight technique for capturing hierarchical relations between fashion labels. In fashion, categories and attributes tend to follow hierarchical relations \cite{papadopoulos2022attentive}. 
HLS shares the predicted labels from the previous stage to the next (labels predicted by GD are shared to CC and its predicted labels are shared with AC). 
For example, ``dress'' is considered ``full-body'' garment while ``shirt'' is an ``upper-body''. Both could have a ``checked'' pattern but only the dress could be classified as a ``pinafore''. We hypothesize that HLS could be instrumental in learning such relations without requiring manual guidance. 
\\ \indent
We examine HLS in two settings: 1) single-task learning (STL) and 2) multi-task learning (MTL).
In STL two separate neural networks are trained for CC and AC while the labels are shared between them with use of embeddings layers as seen in Fig. \ref{fig:hls}. 
The MTL a single neural network is trained for both tasks. We use an encoder-decoder architecture partly inspired by \cite{xu2015show} where encoder produces a representation of the image which is fed into an attention mechanism 
\cite{bahdanau2014neural}. 
The attentive context vector is given to a Gated Recurrent Unit (GRU) along with the embeddings of the garment type. GRU's first hidden state (H1) is used for CC. The context vector is re-calulated based on H1 along with the category embeddings and the same image representation. The produced H2 is used for AC. 


\begin{figure*}[!ht]
    \centering
    \includegraphics[width=1\textwidth]{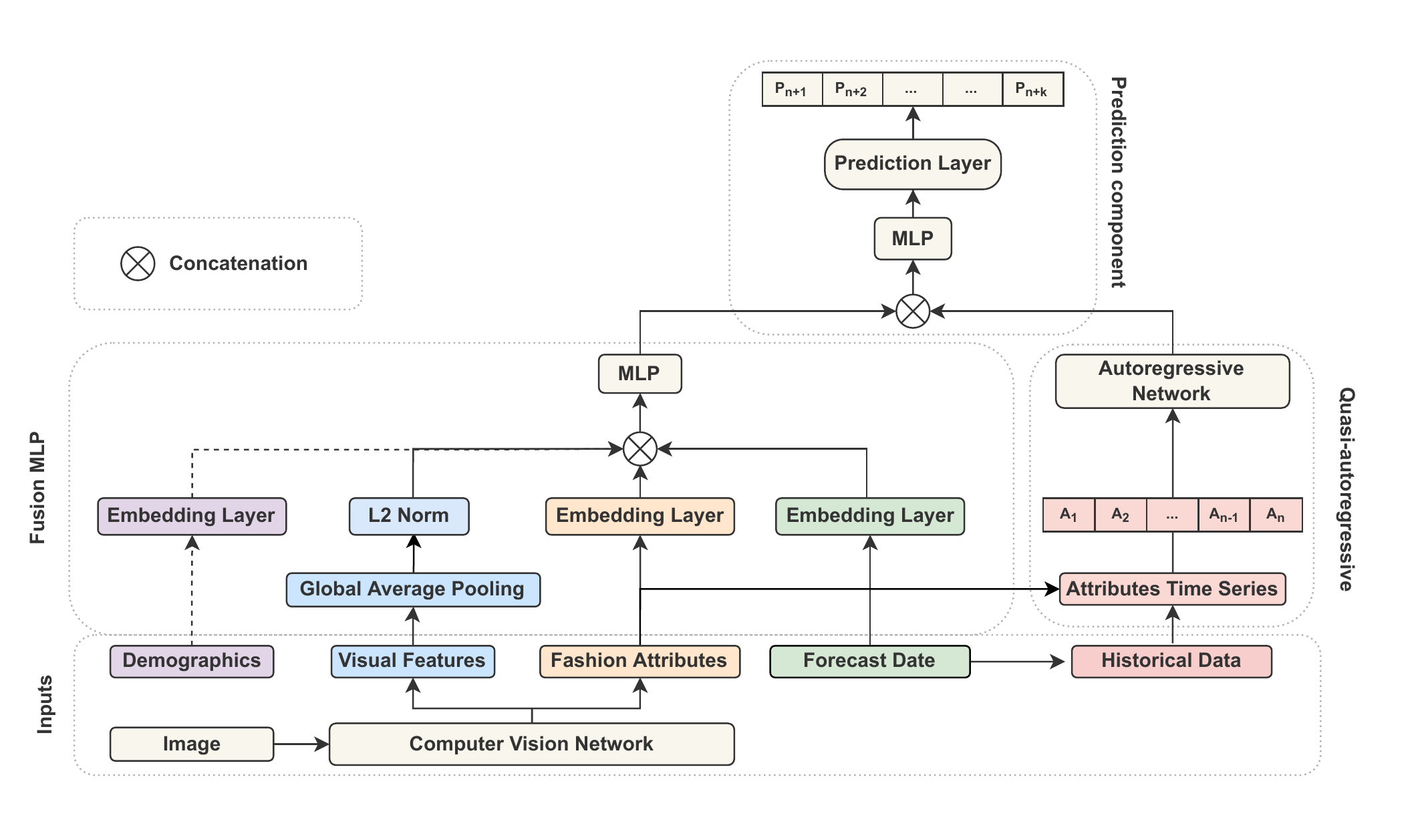}
    \caption{Architecture of MuQAR. Intermittent arrows are optional and applicable only to the Mallzee dataset.}
    \label{fig:muqar}
\end{figure*}

\subsection{New garment popularity forecasting}

New garments by definition lack past data and therefore conventional autoregressive (AR) forecasting models, requiring time series data, can not be effectively utilized.
On the other hand, conventional regression models are not as capable of detecting temporal patterns and dynamics which have a significant role in fashion. 
With this in mind, we utilize MuQAR, a multimodal quasi-autoregressive neural architecture composed of two modules: FusionMLP and quasi-autoregression (QAR). FusionMLP is responsible for representing and combining the multimodal features of a garment; the visual, categorical and temporal aspects \cite{papadopoulos2022muqar}. 
The QAR neural network is modelling the time series of the garment’s category and attributes and we hypothesize that it can be utilized as an informative proxy of temporal dynamics and compensate for the garment's lack of past data.

More specifically, FusionMLP receives (1) the visual features $F_v$ taken from the last convolutional layers of CC and AC models (after applying global average pooling, L2 normalization and concatenating features from CC and AC), (2) the set of predicted fashion category and attributes $a_p$ of garment $p$ and (3) the target forecasting date, in the form of ``day of year'', ``week'', ``month'' and ``season''. 
Another optional input is the target demographic group which is only available in one of the datasets considered here. 
Categorical, temporal and demographic information are represented by distinct learnable embedding layers resulting in $F_c$, $F_t$, $F_g$ of size $d_c$, $d_t$, $d_g$ respectively. 
FusionMLP employs an early fusion approach where $(F_v;F_c;F_t;F_g)$ are concatenated and processed by a multi-layer perceptron (MLP) with $n_{mlp}$ fully connected layers of $u_{mlp}$ RelU activated units resulting into $F_{F}$.

On the other hand, QAR receives the popularity time series $\{A_t\}$ of $a_p$, as predicted by CC and AC. More specifically, it receives the $\textbf{A}=\{A_1,\dots,A_n\}\in\mathbb{R}^{n\times \mid a_p\mid}$ 
that contains $n$ time steps prior to the target date. The outcome of QAR is a vector representation $F_Q\in\mathbb{R}^{q}$ related to the forecast.
MuQAR is a general and flexible architecture since it can incorporate any AR model, whichever is deemed more appropriate for the task. In our study we experiment with a baseline linear regression (LR) and five AR neural networks, widely used for time series forecasting, namely: (1) Long Short Term Memory network (LSTM) \cite{hochreiter1997long}, (2) Feedback LSTM \cite{graves2013generating}, (3) Convolutional Neural Network (CNN) \cite{zhao2017convolutional}, (4) Convolutional LSTM (ConvLSTM) \cite{xingjian2015convolutional} and (5) Transformer \cite{vaswani2017attention}.
Finally, $(F_{F}; F_{Q})$ are concatenated, fed into another MLP and a linear layer predicts the garment's next $k$ popularity time steps $\{P_{n+1},\dots,P_{n+k}\}$ as can be seen in Fig. \ref{fig:muqar}.

\subsection{Integration in the VR application}
In order for the popularity score to be accessible to the end user inside the VR Designer application, we have set up a Google cloud function\footnote{https://cloud.google.com/functions}. This function handles the communication between the VR Designer application and the popularity score estimation service (PSES). In essence, when the end user selects a garment, a request is sent to the function that includes a garment photo thumbnail provided from inside the VR Designer application along with the predefined market segment. The function then propagates the request to 
PSES, and then returns the results to the VR Designer application. The score is shown in the user interface of the app under the Garments section as shown in Fig. \ref{fig:teaser}, under the label "Popularity Score". When the user selects an active garment, then the score is changed accordingly to reflect the current selection. 
Moreover, on the left part of the user interface the user can perform three tasks under each respective field; select an active avatar, garment and animation. There are also controls for avatar visibility, and animation play / pause functionality and speed. On the right of the screen is the assortment section, where a user can save a state of avatar, animation and garment to easier switch between different combinations and create a virtual fashion runway for an external reviewer.


\section{Experimental setup}

\subsection{Fashion image datasets}

\begin{table}
  \centering
  \begin{tabular}{llll}
    \toprule
    \textbf{Dataset} & \textbf{Images} & \textbf{Categories} & \textbf{Attributes} \\ 
    \midrule
    DeepFashion & 289,222 & 46 & 1000 \\
    DeepFashion2 & 491,895 & 13 & - \\
    \midrule
    Mallzee (GD) & 16,550 & 4 & - \\
    Mallzee (CC+AC) & 310,753 & 16 & 110 \\
    \bottomrule
  \end{tabular}
  \caption{Details for the fashion image datasets.}
  \label{tab:data_stats}
\end{table}

\begin{table}
  \centering
  \begin{tabular}{llll}
    \toprule
    \textbf{Dataset} & \textbf{Records} & \textbf{Period} 
    & \textbf{Task}\\ 
    \midrule
    VISUELLE & 5577 & 2016 - 2019 & 
    Regression \\
    SHIFT15m & 15,218,721 & 2010 - 2020 & 
    Regression\\
    MLZ-DG & 5,412,193 & 2017 - 2020 & 
    Regression\\
    Amazon Reviews &  3,002,786 & 2009 - 2014 & 
    Classification\\
    \bottomrule
  \end{tabular}
  \caption{Details for the popularity datasets.}
  \label{tab:data_stats_pop}
\end{table}

For training the three computer vision models GD, CC and AC we make use of three large-scale fashion image datasets, DeepFashion (DF1) \cite{liu2016deepfashion} and DeepFashion2 (DF2) \cite{ge2019deepfashion2} as well as the Mallzee datasets \cite{papadopoulos2022attentive} that also contains annotations for footwear that were lacking from both DF1 and DF2. The number of images and annotated classes for each dataset can be seen in Table \ref{tab:data_stats}.
More specifically, for GD we make use of the Mallzee (GD) dataset and the DF2. We re-map the 13 categories of DF2 into upper-body, lower-body, full-body and create a balanced subset of 50,000 objects per garment type.
We do not make use of DF1 for GD because it only has a single annotated garment per image, even if multiple are depicted. 
For training CC and AC tasks we make use of DF1 and Mallzee (CC+AC) dataset. 

\subsection{Fashion popularity datasets}

For training the popularity forecasting models we make use of three fashion datasets, namely VISUELLE \cite{skenderi2021well}, SHIFT15m \cite{kimura2021shift15m} and the Mallzee-demographics dataset (MLZ-DG) \cite{papadopoulos2022muqar}.
Additionally, we experiment with the Amazon Reviews\footnote{\url{https://jmcauley.ucsd.edu/data/amazon/}} dataset and specifically the Home and Kitchen subset, in order to examine the generalizability of the popularity forecasting model to other domains. 
The number of popularity records, time span and task of each dataset can be seen in Table \ref{tab:data_stats_pop}. 
We follow the pre-processing steps from \cite{papadopoulos2022muqar} for all datasets.
SHIFT15m and Amazon provide pre-computed visual features, extracted from pre-trained models on ImageNet. VISUELLE provides the raw images and the authors use a ResNet50 pre-trained on ImageNet to extract visual features. We follow this workflow so as to ensure comparability. The MLZ-DG does not come with pre-computed visual features, therefore we use the computer vision models presented in section \ref{subsec:cv} in order to extract the fashion labels and visual features. 
It is the only dataset that provides demographic information allowing for more targeted forecasting. It consists of two gender groups (men, women) and 7 age-groups including $<$18, 18-25, 25-30, 30-35, 35-45, 45-55, $>$55. 
For all datasets, expect VISUELLE, 
we compute the mean popularity for each fashion attribute per week in order to create the time series that feed QAR. 

\subsection{Evaluation Protocol}

For the evaluation of the GD task we rely on mean average precision (mAP) averaged over 10 intersection over union thresholds (IoU) (from 0.5 to 0.95 with steps of 0.05 size) and the average Recall@K (AR@K) that signify the average recall given K detections per image. 
For CC we rely on top-K accuracy for K=3,5 and the recall rate at top-K for K=3,5 for AC; since these metrics are also used to benchmark DF1 \cite{liu2016deepfashion}. 
For experiments on the Mallzee (CC+AC) dataset we also report the metrics for K=1.
We split the fashion image datasets into the same training, validation and testing sets as other works so as to ensure a fair comparison. 

For the evaluation of NGPF we use multiple evaluation metrics. For regression tasks we use the: Mean Absolute Error (MAE), Pearson Correlation Coefficient (PCC) and Binary Accuracy (BA) while for classification tasks we used: Accuracy and the Area under the ROC Curve (AUC). We selected the best performances of each model with the use of TOPSIS, a method for multi-criteria decision analysis \cite{hwang1981methods}. 
We sort the forecasting datasets chronologically, separate items with numerous records in the training set and items with only one record in the validation and testing sets. 
However, for VISUELLE we follow the experimental protocol described in \cite{skenderi2021well} to ensure comparability with Weighted Absolute Percentage Error (WAPE) and Mean Absolute Error (MAE) as the evaluation metrics. 

\section{Results}

\subsection{Computer Vision tasks}

\begin{table*}
\centering
  \begin{tabular}{c|c|cccc}
    \toprule
     & Dataset & Faster R-CNN & EfficientDet-D1 & EfficientDet-D2 & CenterNet \\

    \midrule

    mAP & DeepFashion2 & 73.0 & 72.1 & 64.8 & \textbf{75.6} \\ 
    AR@100 &  & 83.7 & 81.9 & 76.7 & \textbf{86.9} \\ 
    \midrule
    mAP & \textit{Mallzee (GD)}
    & \textbf{80.6} & 62.6 & 60.5 & 73.2 \\ 
    AR@100 & & \textbf{85.8} & 75.2 & 70.7 & 80.7 \\     
    \bottomrule
  \end{tabular}
    \caption{Object detection models trained on DF2 and the Mallzee (GD) dataset for garment detection. Evaluations performed in terms of mean Average Precision (mAP) and Average Recall at 100 (AR@100). 
  The DF2 includes three classes (upper, lower, fullbody) while the Mallzee GD also includes footwear. (Bold denotes the best performing model by metric)}
  \label{tab:OD}
\end{table*}

\begin{table}
\centering
  \begin{tabular}{lcccccc}
    \toprule
\textbf{Method} 
& \multicolumn{3}{c}{\textbf{Category}} 
& \multicolumn{3}{c}{\textbf{Attributes}} \\

    \midrule
    & Top-1 & Top-3 & Top-5 & Top-1 & Top-3 & Top-5 \\ 
    \midrule
    
    STL w/o HLS & 90.10 & 98.55 & 99.48 
    & \textbf{78.75} & \textbf{94.32} & 96.78 \\
    
    STL w/ HLS & \textbf{90.67} & \textbf{98.72} & \textbf{99.53}
    & 77.62 & 93.70 & \textbf{96.80} \\
    
    MTL w/ HLS & 87.63 & 98.01 & 99.31 
    & 62.74 & 83.81 & 90.32 \\
    \bottomrule
  \end{tabular}
    \caption{Evaluating hierarchical label sharing for category and attribute classification on the Mallzee (CC+AC) in two settings: STL and MTL. (Bold denotes the best performance)}
      \label{tab:MLZ}
\end{table}

Regarding GD models, we can see in Table \ref{tab:OD} that CenterNet consistently outperforms the other models on DF2; scoring 75.6\% and 86.9\% in terms of mAP and AR@100 respectively, followed by Faster R-CNN.
On the other hand, Faster R-CNN yields the highest scores on the Mallzee (GD) dataset, with 80.6\% mAP and 85.8\% AR@100 for 4 classes while CenterNet comes second. 

As illustrated in Table \ref{tab:MLZ}, for CC on the Mallzee (CC+AC) dataset, STL w/ HLS outperforms the other methods in terms of top-1, top-3 and top-5 accuracy on categories. For AC, HLS does not further improve STL's performance with the exception of a negligible +0.02\% in terms of top-5 recall for attributes. MTL w/ HLS has the lowest performance of the three settings across all metrics.
On the DeepFashion dataset we compare our models against numerous relevant studies as shown in Table \ref{tab:DF1}.
All three models surpass the SotA on CC. Specifically, ``STL w/ HLS'' outperforms its baseline and sets the highest top-3 accuracy with 93.99\% (+0.98\%) while ``MTL w/ HLS'' achieves the highest top-5 accuracy with 97.57\% (+0.56\%).
Additionally, STL w/ HLS improves upon the performance of its baseline STL w/o HLS for AC and surpasses the current SoTA by 6.36\% in terms top-3 recall rate.
Although MTL w/ STL performs very well for CC, we observe that is has a restricted performance on AC. We hypothesize that while the two tasks are related they require different levels of granularity - since fashion attributes are more fine-grained related to textures, fabrics and styles - that may be better captured by separate models.   

\subsection{New Garment Popularity Forecasting}

\begin{table}
\centering
  \begin{tabular}{lcccc}
    \toprule
\textbf{Method} & \multicolumn{2}{c}{\textbf{Category}} & \multicolumn{2}{c}{\textbf{Attributes}} \\
\midrule
    & Top-3 & Top-5 & Top-3 & Top-5 \\ 
    \midrule
    \cite{chen2012describing} & 43.73 & 66.26 & 27.46 & 35.37 \\
    \cite{huang2015cross} & 59.48 & 79.58 & 42.35 & 51.95 \\
    \cite{liu2016deepfashion} & 82.58 & 90.17 & 45.52 & 54.61 \\ 
    \cite{corbiere2017leveraging} & 86.30 & 92.80 & 23.10 & 30.40 \\
    \cite{wang2018attentive} & 90.99 & 95.78 & 51.53 & 60.95 \\
    \cite{ye2019hard} & 90.06 & 95.04 & 52.82 & 62.49 \\
    \cite{li2019two} & 93.01 & 97.01 & 59.83 & \textbf{77.91} \\
    STL w/o HLS & 93.71 & 97.40 & 65.79 & 73.57 \\
    STL w/ HLS & \textbf{93.99} & 97.49 & \textbf{66.19} & 73.73 \\
    MTL w/ HLS & 93.72 & \textbf{97.57} & 53.01 & 66.4\\
    \bottomrule
  \end{tabular}
    \caption{
  Benchmarking on DeepFashion for category and attribute classification. (Bold denotes best performance per metric)
  }
  \label{tab:DF1}
\end{table}

\begin{sidewaystable}
\centering
  \begin{tabular}{ccc|cc|cc|ccc|c}
    \toprule
     \multicolumn{1}{c}{\textbf{Input}} & 
     \multicolumn{1}{c}{\textbf{Rank}} & 
     \multicolumn{1}{c}{\textbf{Model}} & 
     \multicolumn{2}{c}{\textbf{MAE($\downarrow$) }} & 
     \multicolumn{2}{c}{\textbf{PCC($\uparrow$) }} & 
     \multicolumn{3}{c}{\textbf{Accuracy($\uparrow$) }} &
     \multicolumn{1}{c}{\textbf{AUC($\uparrow$) }} 
     \\
     
    \midrule
    
    &&& MLZ-DG & SHIFT15m & MLZ-DG & SHIFT15m & MLZ-DG & SHIFT15m & Amazon & Amazon\\
    
    \midrule
    
    &9& LR & 0.1878 & 0.1162 & 0.2439 & 0.3177 & 63.10 & 59.58 & 48.52 & 65.41 \\
    
    &8& Feedback LSTM & 0.1809 & 0.1149 & 0.3395 & 0.3376 & 64.62 & 61.43 & 44.95 & 67.89 \\
    
    Time&7& Transformer & 0.1842 & 0.1149 & 0.3071 & 0.3398 & 63.67 & 61.28 & \underline{51.10} & \underline{71.29} \\
    
    Series&5& LSTM & 0.1656 & 0.1150 & 0.5109 & 0.3371 & 69.67 & 61.42 & 45.58 & 67.54 \\
    
    (QAR)&4& ConvLSTM & 0.1641 & \underline{0.1147} & 0.5225 & \underline{0.3411} & 69.98 & \underline{61.58} & 46.58 & 68.59 \\
    
    &3& CNN & \underline{0.1611} & 0.1148 & \underline{0.5379} & 0.3406 & \underline{70.99} & 61.51 & 47.18 & 69.34 \\
    \midrule
    
    Visual&6& LR & 0.1599 & 0.1186 & 0.5314 & 0.1940 & 71.86 & 57.93 & 41.46 & 68.18 \\
    
    Features&2& FusionMLP & 0.1074 & 0.1148 & 0.7893 & 0.2811 & 81.52 & 60.89 & 46.96 & 71.69 \\
    
    \midrule
    
    Hybrid&1& MuQAR & \textbf{0.0949} & \textbf{0.1100} & \textbf{0.8362} & \textbf{0.3934} & \textbf{83.41} & \textbf{63.57} & \textbf{51.51} & \textbf{74.24} \\

    \bottomrule
  \end{tabular}
    \caption{Ablation analysis of MuQAR and its modules on two fashion datasets: Mallzee-demographics (MLZ-DG) and SHIFT15m and the Amazon Reviews: Home and Kitchen dataset with 12 weeks as input and 1 as output. The models are ranked by TOPSIS - from worst (9) to best (1) - based on their overall performance. 
  \textbf{Bold} denotes the best overall performance per metric and dataset. 
  \underline{Underline} denotes the best performing QAR network per dataset. 
  }
  \label{tab:results_ablation}
\end{sidewaystable}

\begin{table}
\centering
  \begin{tabular}{l|c|cccc}
    \toprule
     \multicolumn{1}{c}{\textbf{Method}} & 
     \multicolumn{1}{c}{\textbf{Input}} & 
     \multicolumn{2}{c}{\textbf{IN:52, OUT:6}} 
     \\
     \midrule
     && WAPE($\downarrow$) & MAE($\downarrow$) 
     \\
    \midrule
    GTM-Transformer \cite{skenderi2021well} & [T] & 62.6 & 34.2 
    \\    
    Attribute KNN \cite{ekambaram2020attention} & [T] & 59.8 & 32.7 
    \\
    FusionMLP & [T] & \underline{55.15} & \underline{30.12} 
    \\
    \midrule
    
    Image KNN \cite{ekambaram2020attention} & [I] & 62.2 & 34 
    \\
    GTM-Transformer \cite{skenderi2021well} & [I] & 56.4 & 30.8 
    \\
    FusionMLP & [I] & \underline{54.59} & \underline{29.82} 
    \\
    
    \midrule

    
    QAR: Transformer & [G] & 62.5 & 34.1 
    \\
    QAR: LSTM & [G] & 58.7 & 32.0 
    \\
    QAR: ConvLSTM & [G] & 58.6 & 32.0 
    \\    
    
    GTM-Transformer \cite{skenderi2021well} & [G] & 58.2 & 31.8 
    \\
    
    QAR: Feedback LSTM & [G] & 58.0 & 31.7 
    \\
    QAR: CNN & [G] & \underline{57.4} & \underline{31.4} 
    \\
    
    \midrule
    
    Attribute + Image KNN  \cite{ekambaram2020attention} & [T+I] & 61.3 & 33.5 
    \\
    Cross-Attention RNN \cite{ekambaram2020attention} & [T+I] & 59.5 & 32.3 
    \\
    GTM-Transformer \cite{skenderi2021well} & [T+I] & 56.7 & 30.9 
    \\    
    FusionMLP & [T+I] & \underline{54.11} & \underline{29.56} 
    \\    
    
    \midrule

    GTM-Transformer AR \cite{skenderi2021well} & [T+I+G] & 59.6 & 32.5 
    \\    
    
    Cross-Attention RNN+G \cite{ekambaram2020attention} & [T+I+G] & 59.0 & 32.1 
    \\
    
    GTM-Transformer \cite{skenderi2021well} & [T+I+G] & 55.2 & 30.2 
    \\    

    MuQAR w/ Transformer & [T+I+G] & 54.87 & 29.97 
    \\
    MuQAR w/ Feedback LSTM & [T+I+G] & 54.37 & 29.7 
    \\    
    MuQAR w/ LSTM & [T+I+G] & 54.3 & 29.66 
    \\

    MuQAR w/ CNN & [T+I+G] & 53.9 & 29.44 
    \\
    MuQAR w/ ConvLSTM & [T+I+G] & \underline{53.61} & \underline{29.28} 
    \\
    
    \bottomrule
  \end{tabular}
    \caption{Comparative analysis between MuQAR and its modules against SotA forecasting models on the VISUELLE dataset using 52 week-long time series as input from Google Trends and forecasting the next 6. Features used: [T]ext, [I]mage, [G]oogle trends. 
  \underline{Underline} denotes the best performing network per input. 
  }
  \label{tab:results_visuelle}
\end{table}

\begin{figure}[!t]
    \centering
    \includegraphics[width=0.7\textwidth]{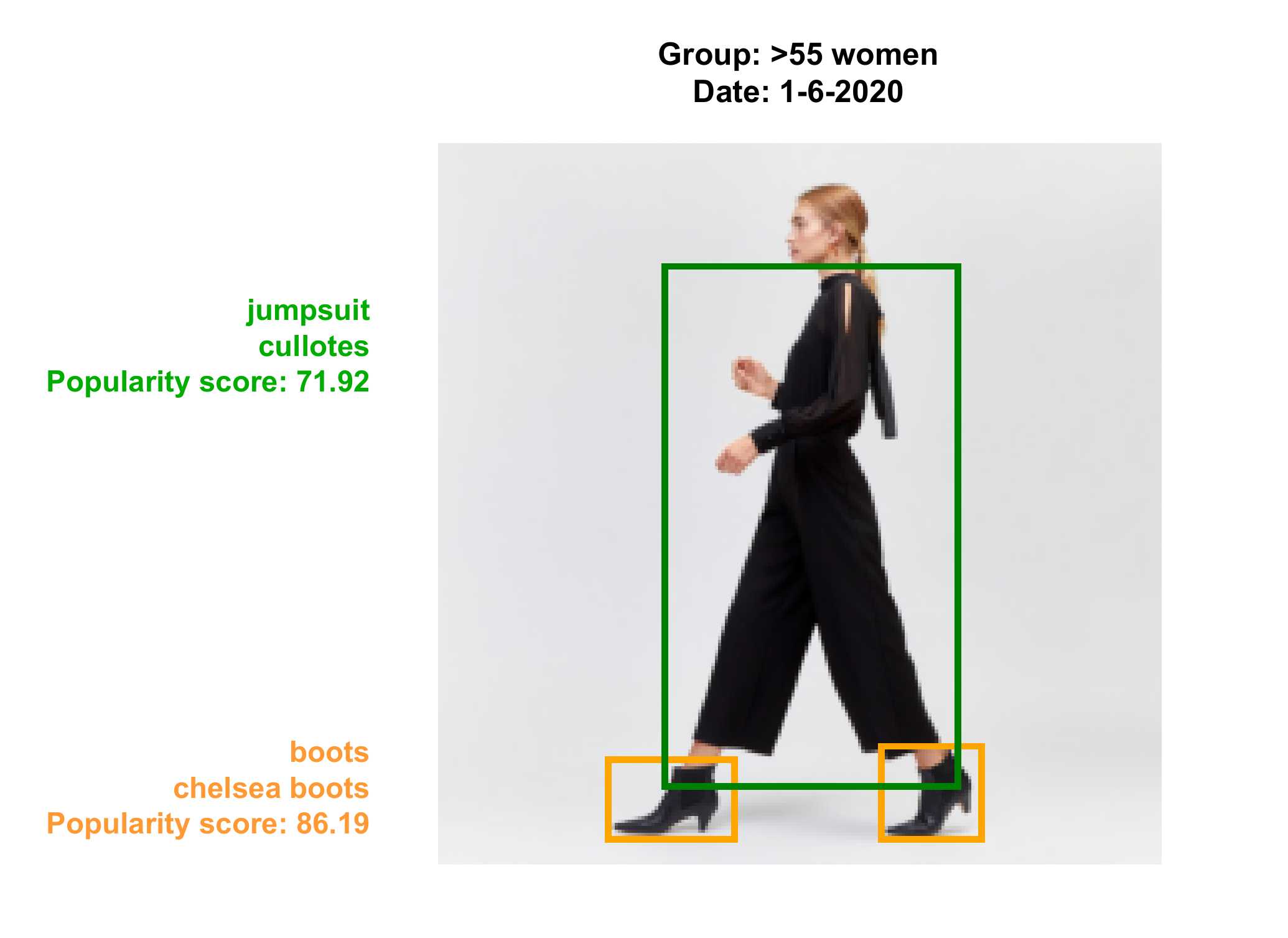}
    \caption{Inference sample by MuQAR}
    \label{fig:inference}
\end{figure}

For NGPF, we first perform an ablation analysis on the MLZ-DG, SHIFT15m and Amazon Reviews comparing the MuQAR against its modules; FusionMLP and various QAR models. 
Firstly, we can observe in Table \ref{tab:results_ablation} that the CNN, ConvLSTM and Transformer QAR networks yield the best performance for MLZ-DG, SHIFT15m and Amazon datasets respectively.
We integrate these specific QAR models in the MuQAR experiments for the three datasets.
Secondly, FusionMLP seems to outperform the QAR models on MLZ-DG but not on SHIFT15m and Amazon Reviews.
The visual features in MLZ-DG are extracted by a model fined-tuned on fashion imagery, while the other provide visual features from pre-trained models on ImageNet, thus are specialised on fashion imagery. The quality and specialisation of the visual features should play an important role for the given task and may have caused the restricted performance of FusionMLP. 
Finally, we observe that MuQAR is capable of consistently outperforming QAR models and FusionMLP on all the three datasets. Combining the visual features (even if they are not fine-tuned for fashion) with the popularity time series of the categorical labels, MuQAR is able to improve upon the task of NGPF.
In Fig. \ref{fig:inference} we present an inference example by MuQAR.

Additionally, we perform a comparative analysis on the VISUELLE dataset between MuQAR and multiple SotA models presented in Table \ref{tab:results_visuelle}. 
We observe that w/ ConvLSTM has the highest overall performance. Not only that, but MuQAR with any QAR model consistently surpasses all other models.
Surprisingly, our FusionMLP when utilizing only categorical labels and images [T+I] or only images [I] can outperform the GTM-Transformer and Cross-Attention RNN that also employ time series [T+I+G]. 
Moreover, QAR:CNN and QAR:Feedback LSTM can outperform GTM-Transformer when only using time series [G] while being simpler and more lightweight architectures.

\section{Discussion}

In this study we address the task of forecasting the popularity of new garment designs based on visual features. Our objective is to provide trends insights to fashion designers within an interactive VR application. 
This is a challenging task for two main reasons. The forecast model should be supplementary and non intrusive to the design process. Therefore it has to rely solely on the visual aspects of a garment and not require additional input from the designer. 
Additionally, new products by definition lack past data while fashion trends are 
constantly evolving.

To address the first challenge, we utilize a computer vision pipeline for feature extraction and classification on fashion imagery.
It utilizes a hierarchical label sharing technique that captures hierarchical relations between fashion categories and attributes. It surpasses the SotA on the DeepFashion benchmark dataset with 93.99\% top-3 accuracy and 97.57\% top-5 accuracy for category and 66.19\% top-3 recall rate for attribute classification. 

To address the second challenge, we utilize a multimodal quasi-autoregressive neural network, MuQAR, that utilizes the visual aspects of a garment along with the popularity time series of the garment's category and attributes;
the latter working as a proxy for the garments lack of past data.
An ablation study on three dataset proves the validity of the proposed method 
while also surpassing SotA models, by +2.88\% improvement in terms of WAPE and +3.04\% in terms of MAE on the VISUELLE dataset.

We have deployed the aforementioned models in API endpoints and have integrated them in a VR application for fashion designers, even though it could just as well be utilized by conventional design apps. 
By providing interactive forecasts we hope to facilitate the creative process of fashion designers and offer the choice to fine tune their designs based on current consumer preferences.
This could potentially translate in higher profits for the brand and also play a part in mitigating the problem of unsold inventory which contributes to the industry's high environmental impact. 


\section{Acknowledgments}
This work is partially funded by the project ``eTryOn - virtual try-ons of garments enabling novel human fashion interactions'' under grant agreement no. 951908. The authors would also like to thank Jamie Sutherland, Manjunath Sudheer and the company Mallzee/This Is Unfolded for the data acquisition as well as all the useful insights and feedback.

\bibliographystyle{unsrt}  
\bibliography{references}

\end{document}